\documentclass[10pt,twocolumn,letterpaper]{article}

\usepackage{iccv}
\usepackage{times}
\usepackage{epsfig}
\usepackage{graphicx}
\usepackage{amsmath}
\usepackage{amssymb}
\usepackage{pifont}
\usepackage{booktabs}
\usepackage{multirow}
\usepackage{arydshln}
\usepackage{lipsum}  
\usepackage{enumitem}
\usepackage[table]{xcolor}
\usepackage{bbm}

\newcommand{\xmark}{\ding{55}}%

\def\0{{\bf 0}}
\def\1{{\bf 1}}

\newcommand*{\mathcolor}{}
\def\mathcolor#1#{\mathcoloraux{#1}}
\newcommand*{\mathcoloraux}[3]{%
  \protect\leavevmode
  \begingroup
    \color#1{#2}#3%
  \endgroup
}

\def\etal{{\em et al.\/}\,}
\def\eg{{\em \textit{e.g.}\/}\,}
\def\ie{{\em \textit{i.e.}\/}\,}

\def\distillvlm{{DistillV{\small L}M}\,}

\usepackage{listings}
\usepackage{subcaption}

\definecolor{darkgreen}{rgb}{0.0, 0.26, 0.15}
\definecolor{maroon}{rgb}{0.76, 0.13, 0.28}

\definecolor{codegreen}{rgb}{0,0.6,0}
\definecolor{codegray}{rgb}{.95,.95, .95}
\definecolor{codepurple}{rgb}{0.58,0,0.82}
\definecolor{backcolour}{rgb}{0.95,0.95,0.92}
\definecolor{p}{rgb}{148,2,209}

\usepackage{xcolor}
\usepackage[pagebackref=true,breaklinks=true,colorlinks,bookmarks=false]{hyperref}
\hypersetup{
colorlinks=true,
linkcolor=red,
filecolor=magenta,      
urlcolor=cyan,
}

\iccvfinalcopy 

\ificcvfinal\fi
\begin{document}

\title{Compressing Visual-linguistic Model via Knowledge Distillation}

\author{  Zhiyuan Fang$^\dag$ , Jianfeng Wang$^\ddag$, Xiaowei Hu$^\ddag$, Lijuan Wang$^\ddag$,  {Yezhou Yang}$^\dag$, {Zicheng Liu}$^\ddag$\\   $^\dag$Arizona State University, \ \ \ \ \ \ \ \ \ \ \ \ \ \ \ \ $^\ddag$Microsoft Corporation  \\
 {\small \texttt{\{zy.fang, yz.yang\}@asu.edu}, \ \ \ \ \ \ \ \ \ \  \texttt{\{jianfw, lijuanw, leizhang, zliu\}@microsoft.com}} \\
}

\maketitle

\begin{abstract}
Despite exciting progress in pre-training for visual-linguistic (VL) representations, very few aspire to a small VL model.  In this paper, we study knowledge distillation (KD) to effectively compress a  transformer based large VL model into a small VL model. The major challenge arises from the inconsistent regional visual tokens extracted from different detectors of Teacher and Student, resulting in the misalignment of hidden representations and attention distributions. To address the problem, we retrain and adapt the Teacher by using the same region proposals from Student's detector while the features are from Teacher's own object detector. 
With aligned network inputs, the adapted Teacher is capable of transferring the knowledge through the intermediate representations.
Specifically, we use the mean square error loss to mimic the attention distribution inside the transformer block, and present
a token-wise noise contrastive loss to align the hidden state
by contrasting with negative representations stored in a sample queue. 
To this end, we show that our proposed distillation significantly improves the performance of small VL models on image captioning and visual question answering tasks.
It reaches $120.8$ in CIDEr score on COCO captioning, an improvement of $5.1$ over its non-distilled counterpart; and an accuracy of $69.8$ on VQA 2.0, a $0.8$ gain from the baseline.
Our extensive experiments and ablations confirm the effectiveness of VL distillation in both pre-training and fine-tuning stages.

\end{abstract}

\section{Introduction}
There have been exciting progress in visual linguistic~(VL) pre-training to learn omni-representation models~\cite{lu2019vilbert,su2019vl,chen2019uniter,tan2019lxmert,zhou2020unified,li2020oscar} which could benefit a number of downstream tasks (\ie, image captioning, VQA, image retrieval, etc.). The success can largely be attributed to the self-attention-based~\cite{vaswani2017attention} transformer architecture, \eg, BERT~\cite{devlin2018bert}, which is effective in learning from image-text pairs at scale. So far, much of the work has focused on large models that suffer from high latency and large memory footprints at the time of inference, which limits their deployment to resource constrained edge devices for real-world applications.

\begin{figure}[t!]
    \centering
    \includegraphics[width=.48\textwidth]{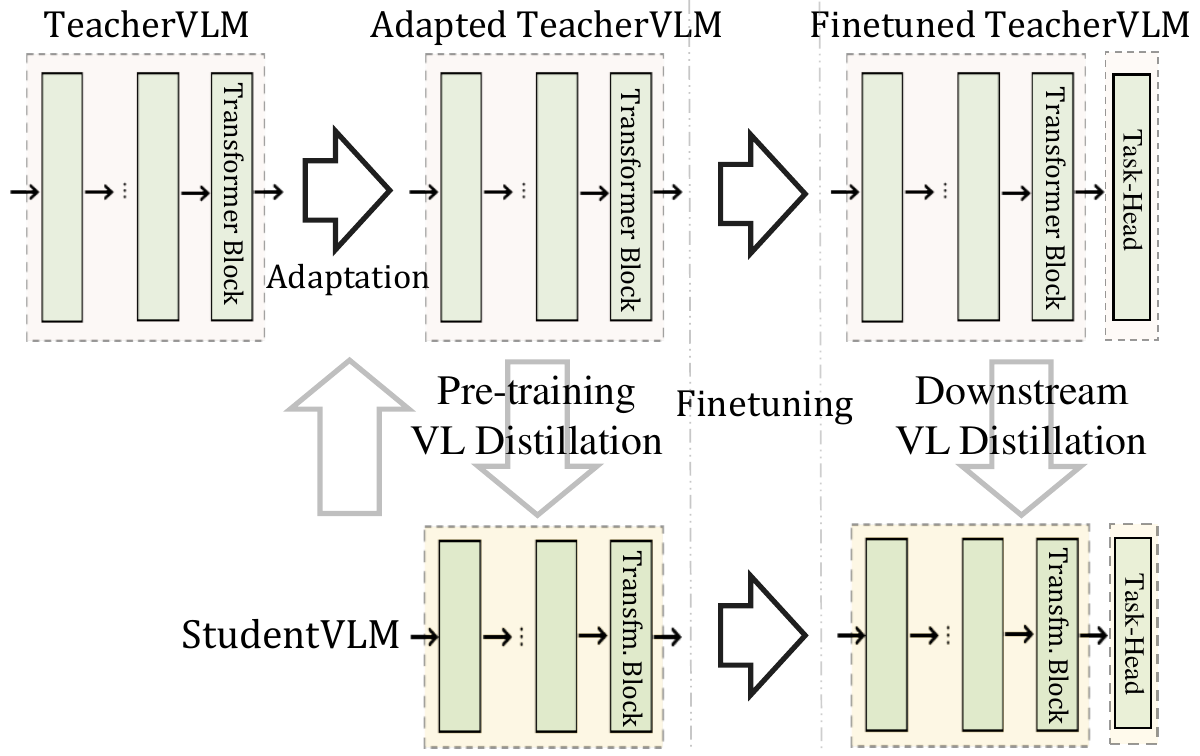} 
    \caption{\small Overview of our proposed VL distillation schema. 
    The VL model typically contains a region feature extraction module and a multi-modal transformer module. To have an aligned input, we adapt the Teacher VL model based on the region proposals from Student's region feature extractor. The VL distillation is then performed in both the pre-training stage and the fine-tuning stage.
    }
    \label{fig:abstract}
    \vspace{-2mm}
\end{figure}

As one of the effective techniques to compress large models, 
knowledge distillation (KD)~\cite{hinton2015distilling, bucilua2006model} was proposed  by injecting the knowledge from a strong Teacher model into a smaller Student model without losing too much generalization power. 
Typically, the knowledge is transferred though
mimicking the output logit~\cite{hinton2015distilling,sanh2019distilbert,fang2021seed}, reducing the divergence of feature maps~\cite{zagoruyko2016paying,NST2017,yim2017gift}, or learning the intermediate layer representations~\cite{koratana2019lit,ahn2019variational}, \etc. 

In recent years, KD has been proven effective in compressing language models. For instance, Kim \etal~\cite{kimrush2016sequence} adopt KD for sequential model compression. In the transformer based language model, DistillBERT~\cite{sanh2019distilbert} reduces the size of the BERT-base model by 40\% using a cosine embedding loss on the basis of hidden embedding in the transformer block, and a soft-target probability loss. TinyBERT~\cite{jiaoetal2020tinybert}, MobileBERT~\cite{sun2020mobilebert} and MiniLM~\cite{wang2020minilm} further highlight the importance of minimizing the self-attention distributions across Teacher and Student networks. In particular,~\cite{clark2019does} visually shows that attention maps in BERT capture substantial linguistic knowledge and syntactic relations that provide critical information during the distillation~\cite{jiaoetal2020tinybert}.

Heretofore, these advances have not been carried over to VL model compression.
We identify the major challenges that prevent us from applying these techniques directly to VL distillation:
Most existing VLP works~\cite{zhou2020unified,li2020oscar} use pre-trained object detector (\eg, Faster-RCNN~\cite{ren2015faster}) to extract regional features as visual tokens then feed them into the multi-modal transformer network for VL pre-training.
A smaller VL model usually uses a lightweight detector for faster inference (\eg, EfficientNet~\cite{tan2019efficientnet} based detector is adopted in~\cite{wang2020minivlm} as visual feature extractor) that may be different from Teacher's detector. The object proposals from the two different detectors are usually very different, and there is no easy way to obtain the semantic correspondence between the two sets of object proposals. It is therefore unable to align the attention distributions or hidden embeddings between Student and Teacher.

\begin{figure*}[th!]
    \centering
    \includegraphics[width=.98\textwidth]{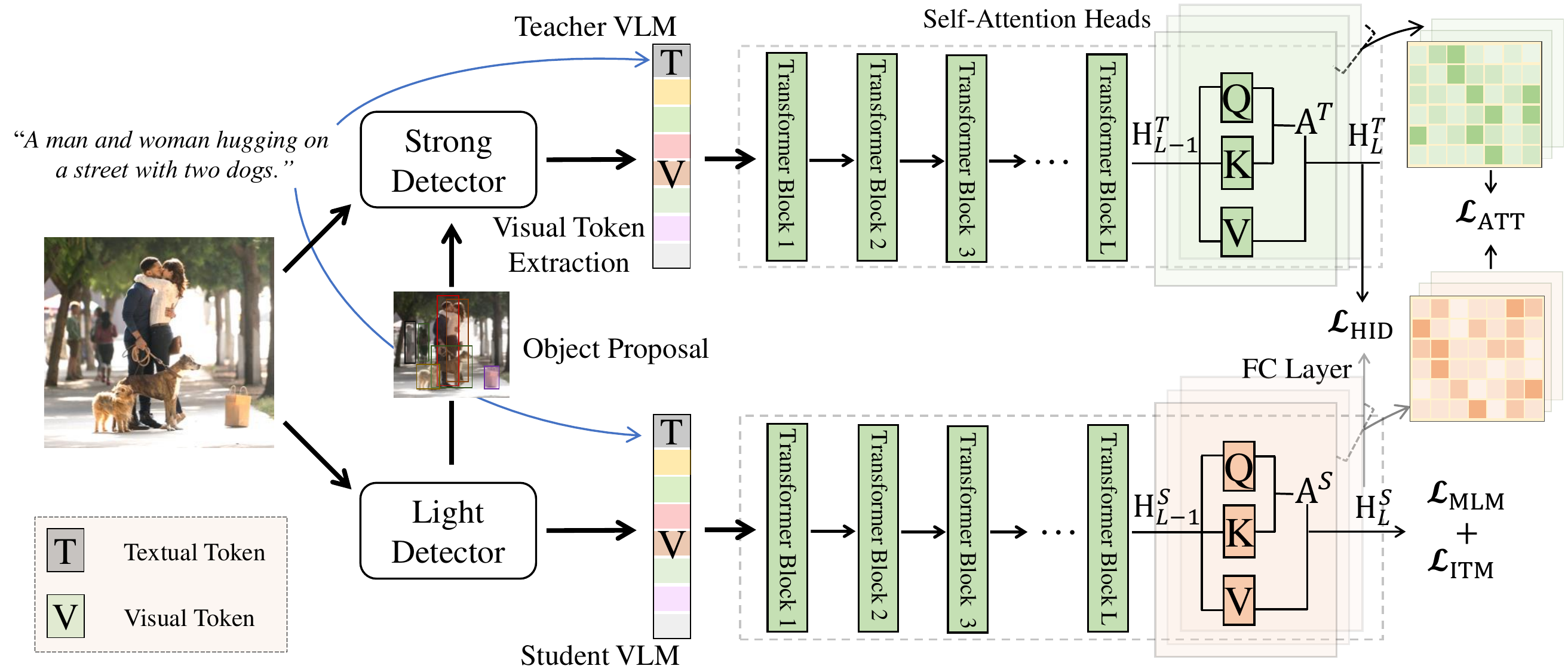} 
    \caption{ 
    {
    Illustration of our proposed DistillV{\footnotesize L}M architecture. The lightweight detector extracts the region features, and the region proposals are injected into the strong detector so that the region features are aligned between Teacher and Student. The Teacher transformer network is adapted with the new input before distillation. The Student VLM is distilled based on the hidden embedding matching and attention distribution alignment. 
    }
    }
    \label{fig:arch}
\end{figure*}

To address the aforementioned challenges, we propose a set of strategies to enable distillation of VL models. First, instead of using object proposals from two different detectors, we use the same set of object proposals, obtained from Student's lightweight detector for the visual token extraction of both Teacher and Student (as shown in Figure~\ref{fig:arch}). This ensures the semantic correspondence between the Teacher and Student's visual tokens. Second, we use a loss term to have the Student to mimic the Teacher's self-attention distribution at the last transformer layer. Third, We further distill the knowledge from the outputs of the transformer layers (\ie, the hidden embeddings). We find that simply learning from the layer-wise Teacher embedding does not provide adequate supervision for the  distillation. Hence, we use a noise contrastive loss to align the token embeddings by contrasting them with randomly sampled negative embeddings that are held in a sample queue. 
Figure~\ref{fig:abstract} gives an overview of our proposed VL distillation schema, where VL distillation is applied for both the pre-training and fine-tuning stages.
In order to examine the effectiveness of our VL distillation, we 
choose the same compact transformer architecture used in~\cite{wang2020minilm,wang2020minivlm}, and the lightweight object detector as in~\cite{wang2020minivlm}, but leverages knowledge distillation techniques to facilitate the training of the small VL model (dubbed as \distillvlm). 
We show that our \distillvlm achieves a comparable performance to a large VL model, and clearly outperforms its non-distilled counterpart~\cite{wang2020minivlm}.

To summarize our contributions:  \\ [-2.5ex]
\begin{itemize}[noitemsep,nosep]
    \item For the first time, we propose VL distillation, a technique that leverages knowledge distillation to facilitate training of smaller VL models. \\ [-1.7ex]
    \item Compared to non-distilled VL model pre-training, VL distillation offers a significant boosting in performance for VL tasks such as image captioning and visual question answering: DistillV{\small L}M achieves 120.8 in  CIDEr score on COCO captioning~\cite{lin2014microsoft} and 69.8 in accuracy on VQA~\cite{balanced_vqa_v2} tasks, which are $5.1$ more or $0.8$ higher than the VL pre-training baselines.\\ [-1.7ex]
    \item We provide extensive ablations of \distillvlm, and systematically analyze the effect of various KD strategies.
    This provides insights for future research on VL model distillation.
\end{itemize}

\section{Related Work}
\noindent\textbf{Visual-linguistic Pre-training.} Following the prominent progress in the transformer-based~\cite{vaswani2017attention} pre-training in natural language~\cite{devlin2018bert,radford2018improving,lagler2013gpt2,brown2020language,clark2020electra,raffel2019exploring}, visual-linguistic pre-training models, either for image+text~\cite{lu2019vilbert,tan2019lxmert,chen2019uniter,li2020oscar,hu2020vivo,zhang2021vinvl,li2020closer,gan2020large,li2020hero,lu202012} or for video+text~\cite{sun2019videobert,li2020hero,miech2020end,zhu2020actbert,lei2021less}, have achieved great success on a number of downstream V+L tasks. Most existing VL models are designed in a two-step fashion: a pre-trained object detector is used to encode the image as
set of regional features (as offline visual tokens) followed by pre-training on a large scale visual-linguistic corpus using tasks like masked language modeling, image-text matching or masked region modeling losses. In particular, Zhang \etal~\cite{zhang2021vinvl} demonstrate the significant role of visual features in VL pre-training and looks for more effective visual representations from a larger object detector. Li~\etal~\cite{li2020oscar} shows that a larger transformer VL model can learn better from larger VL corpus. However, the marginal costs are greater than the marginal benefits. 
Recently, Wang \etal~\cite{wang2020minivlm} propose a small VL model called MiniVLM that uses a lightweight visual feature extractor and smaller transformer to reduce the model size by 73\% and maintain good accuracy on VL tasks. 
Nevertheless, the cost of pre-training on MiniVLM is associated with sub-optimal efficiency: it requires a large amount of training data (14M) to learn a good representation. Thus, it is worth exploring a more efficient way to train small VL models. 
There are other lines of VL pre-training works in which grid features~\cite{huang2020pixel,jiang2020defense} are extracted from the convolutional layers without the proposal computation.~\cite{ramesh2021zero,radford2021learning,desai2020virtex} learn visual representation from scratch using Convolutional Neural Network as image encoder with a transformer for VL pre-training on a large amount of image-text pairs.  The notion of VL distillation is not limited to just the two-stage VL models, it can potentially benefit other types of transformer based VL models as well. \\[-1.5ex]

\noindent\textbf{Knowledge Distillation} has been applied to model compression task across different domains with its main goal being to transfer the ``\textit{knowledge}'' $f(x_i)$ of sample ($x_i, y_i$) from a strong Teacher network ($T$) to the Student network ($S$) by minimizing the divergence between them:
\begin{equation}
    \mathcal{L} = \frac{1}{N}\sum_{i=1}^{N}\bigg(\mathcal{L}_\text{S}(x_i, y_i) + \mathcal{L}_\text{KD}\Big(f^S(x_i), f^T(x_i)\Big)\bigg),
\end{equation}
where $\mathcal{L}_\text{S}(\cdot)$ refers to the original supervision signal(s) on the Student. In practice, this term can possibly be replaced by the exclusive use of $L_\text{KD}$. Depending on the type of knowledge transferred, $\mathcal{L}_\text{KD}$ can derive from soft cross-entropy, mean squared error (MSE) function or \textit{KL}-divergence.
For example,~\cite{hinton2015distilling,bucilua2006model} transfer the learned knowledge by mimicking the mass function of the output probability across classes, or by minimizing the divergence of intermediate features~\cite{yim2017gift,koratana2019lit,huang2017like,yalniz2019billion,xie2020self}.~\cite{tian2019contrastive,tian2019contrastive,fang2021seed} propose contrastive distillation for visual representation learning.  In addition, remarkable advances have been made in knowledge distillation for language model compression (\ie, BERT~\cite{devlin2018bert}), and these works show that mimicking the distribution of self-attention and intermediate representations of transformer blocks increases performances~\cite{sanh2019distilbert,jiao2019tinybert,sun2020mobilebert,xu2020bert} for downstream tasks.
In particular, in the transformer-based language model distillation, DistillBERT~\cite{sanh2019distilbert} proposes to train the small BERT by mimicking the Teacher's output probability of masked language prediction and the embedding features.  TinyBERT~\cite{jiao2019tinybert} and MobileBERT~\cite{sun2020mobilebert} leverage the layer-wise attention distributions for distillation with MSE function.~\cite{wang2020minilm} suggests distilling on the last transformer layer and bringing extra flexibility for training.~\cite{sun2020contrastive,chen2020wasserstein} also use the contrastive distillation in transformer based language model compression.~\cite{fang2021seed,sun2020contrastive} propose using a sample queue to store history embeddings and show that contrasting with more negative samples is beneficial for knowledge distillation.

\section{Visual-linguistic Knowledge Distillation}
Compared to knowledge distillation in language models, VL knowledge distillation requires knowledge transferring from Teacher to Student in both modalities.
We present {DistillV{\small L}M} for the task of visual-linguistic distillation (the overall architecture is illustrated in Figure~\ref{fig:arch}), together with the detailed strategies for our model training. 

\subsection{Visual Token Alignment}
VL pre-training methods such as  OSCAR~\cite{li2020oscar} take as input an image-text pair in the format of Word-Tag-Image triple ($\boldsymbol{w}$, $\boldsymbol{q}$, $\boldsymbol{v}$), where $\boldsymbol{w}$ and $\boldsymbol{q}$ denote the sequence of caption embedding and the word embedding of detected object tags (in texts). 
To obtain the visual tokens $\boldsymbol{v}$ and object tags, a set of image regional vectors are extracted from an object detector. A Faster R-CNN~\cite{NIPS2015_14bfa6bb} detector pre-trained on Visual Genome~\cite{krishna2017visual} is used to extract the visual feature vector of each region, which is concatenated with its regional position coordinates to form a positional-sensitive region feature vector. This vector is then fed into a linear projection to ensure that the final vector $\boldsymbol{v}$ has the same dimension as the caption/tag embedding. The VL pre-training can be seen as a semantic alignment process between the image regions and the textual units. It is worth mentioning that, the top regions to be extracted from the image
is dependent on their associated confidence score output by the detector~\cite{wang2020minivlm}, which leads to some over-sampled and noisy visual tokens. Typically, the order of visual tokens is specified in descending order using the confidence score.
As an alternative to Faster-RCNN, MiniVLM~\cite{wang2020minivlm} uses a lightweight detector (\ie, TEE) in which the backbone is replaced with EfficientNet~\cite{tan2019efficientnet} and a BiFPN~\cite{tan2020efficientdet} module is added to generate multi-scale features. These strategies obviously accelerate inference process, but inevitably also lead to different visual tokens between the Teacher and Student networks during distillation. For this reason, the direct application of the distillation loss to attention matrices or hidden representations leads to an invalid transfer of knowledge. Hence, we extract and align the Teacher/Student's visual tokens by using the same set of detected bounding boxes recognized by the lightweight detector, and keep the same token orders based on their confidence scores (as in Figure~\ref{fig:arch}). Both the Teacher and Student VLM use the same object tags from the lightweight detector during the distillation. Having the Teacher use the visual tokens extracted by proposals from the lightweight detector may result in small performance drop. In practice, we address this issue by fine-tuning/re-training the Teacher VLM using the new visual tokens (Teacher adaptation).

\subsection{Attention Distribution Distillation}
One critical component of the transformer block is the multi-head self-attention module~\cite{vaswani2017attention}.  which enables contextualized information to be captured from an input sequence. A multi-head attention module outputs a set of attended values:
\begin{equation}
    \texttt{Attention(}\mathbf{Q}, \mathbf{K}, \mathbf{V}\text{)}=\texttt{softmax(}\frac{\mathbf{Q}\cdot \mathbf{K}}{\sqrt{d_k}}\text{)}\cdot \mathbf{V},
\end{equation}
where $\mathbf{Q}$, $\mathbf{K}$, and $\mathbf{V}$ denote query, key and value that are retrieved after three independent linear transformations on the hidden embedding $\mathbf{H}_i$ from $i$-th transformer block, and $d_k$ is the dimension of key as a scaling factor. The dot-product between key and query after the softmax normalization is the attention matrix:
\begin{equation}
    \mathbf{A}=\texttt{softmax(}{\mathbf{Q}\cdot \mathbf{K}}/{\sqrt{d_k}}\text{)}.
\end{equation}
Each transformer block consists of a set of consecutive linear transformations, which include one multi-head attention module, a two-layer feed forward network, followed by a normalization layer, and finally a residual connection.

Previous attempts in language model distillation~\cite{jiao2019tinybert, sun2020mobilebert} have demonstrated the importance of transferring self-attention matrices, that are believed to contain latent linguistic information, \eg, syntactic and co-reference relation of input tokens~\cite{clark2019does,jawahar2019does}. 
~\cite{wang2020minilm} shows that using just the last transformer block's attention map yields equivalent results, allowing the Teacher and Student to have a different number of layers. In the case of the VL pre-training task, Cao \etal~\cite{cao2020behind} show that certain attention matrices of the pre-trained VL models contain extensive intra-modal and cross-modal co-reference relations.
These visual-linguistic knowledge is implicitly encoded, but shows a very promising potential for VL distillation.
We formulate the distillation loss of the attention distribution by minimizing the divergence between the self-attention matrices of the last layer of the Teacher and the Student:
\begin{equation}
    \mathcal{L}_\text{ATT} = \frac{1}{T\!\cdot\!H}\sum_{i=1}^{T}\sum_{j=1}^{H} \texttt{MSE}(\mathbf{A}^S_{i, j}, \mathbf{A}^T_{i, j}),
\end{equation}
where $T$, $H$ denote the number of tokens and attention heads in a transformer. $\mathbf{A}_{i, j}$ is the normalized attention for $i$-th token at $j$-th head. 
We further study the effects of the distillation over the attention distribution in ablations.

\subsection{Hidden Representation Distillation}
Similar to previous works~\cite{jiao2019tinybert,sun2020mobilebert}, we also use the hidden representations for the Teacher and Student alignment during distillation. In particular, previous efforts formulate the task as minimizing the divergence of the hidden embedding ($\mathbf{H}\in\mathbb{R}^{T\times d}$) of every Transformer block, whose objective is as follows:
\begin{equation}
    \mathcal{L}_\text{HID-MSE} = \frac{1}{T\!\cdot\!L}\sum_{i=1}^{T}\sum_{j=1}^{L} \texttt{MSE}(\mathbf{H}^S_{i, j}\mathbf{W}_{h}, \mathbf{H}^T_{i, j}),
\end{equation}
and $L$ stands for the number of transformer blocks. $\mathbf{W}_{h}$ is a learnable linear transformation that maps the Student hidden embedding into the identical dimension of Teacher embedding. However, there are limitations for such layer-to-layer alignment method. For example, TinyBERT must employ a uniform-function mapping to selectively choose a subset of the layers for learning, and MobileBERT requires the Teacher and Student to have identical number of layers. Since visual tokens are noisy during the VL distillation, this also leads to an increased difficulty in alignment. Sun \etal~\cite{sun2020contrastive} propose CoDIR, which takes advantage of the noise contrastive estimation (NCE) loss to align the Teacher \& Student's hidden representations by contrasting the target instance ($\mathbf{h}^S$) with more random instances as negative samples and aligning with its positive sample ($\mathbf{h}^T$), $\mathbf{h}\in\mathbb{R}^{d_T}$. Following~\cite{he2020momentum,fang2021seed,sun2020contrastive}, we employ a pre-defined instance queue $[\mathbf{h}_0^T, \mathbf{h}_1^T \cdots \mathbf{h}_K^T]$ to store $K$ random sampled embeddings and one positive embedding from the Teacher network. And the objective of NCE is as:
\begin{equation}
\mathcal{L}_\text{HID} = -\text{log} \frac{\text{exp}(\mathbf{h}^S_i\cdot\mathbf{h}^T_i/\tau)}{\sum_{j=0}^K\text{exp}(\mathbf{h}^S_i\cdot\mathbf{h}_j/\tau)},
\end{equation}
where $\tau$ denotes the temperature hyper-parameter,  $\langle \cdot \rangle$  is the cosine similarity function. There are different ways to retrieve hidden representations $\mathbf{h}$, \eg,~\cite{sun2020contrastive} uses mean-pooled token representations as layer-wise summarized embedding.
We find that applying the NCE loss to token-wise embedding leads to better distillation results, as discussed in Section~\ref{sec:method}. A linear mapping is introduced for the identical dimension transformation: $\phi: \mathbb{R}^{d_S} \rightarrow \mathbb{R}^{d_T}$ ($d_S$, $d_T$ denote the hidden embedding dimension for Student and Teacher networks). 
To update the instance queue, we en-queue
the Teacher-derived representation of the current batch ($\mathbf{h}^T$) and de-queue the earliest stored samples after the iteration. The introduction of the queue design enables batch-size independent distillation and allows the comparison with more contrastive samples with limited computational resources. In ablations, we discuss the effect of enlarging queue size and other distillation methods. In contrast to~\cite{he2020momentum,sun2020contrastive}, we store representations from the pre-trained and frozen Teacher network in the sample queue, which remain constant during training. This frees us from the use of momentum encoder like in~\cite{fang2021seed}.

\subsection{Classification Distillation}
The losses mentioned above allow the task-agnostic distillation during the pre-training stage. In addition, in the fine-tuning stage, we carry out knowledge distillation that benefits certain VL downstream tasks. Specifically, most VL downstream tasks are classification based tasks with labels, \eg, image captioning or VQA tasks.  Continuing the distillation at the downstream alleviates the domain gap brought by different pre-training VL corpus. As in~\cite{hinton2015distilling}, we minimize the softmax prediction of Student and Teacher networks and the loss is measured by the cross-entropy:
\begin{equation}
\mathcal{L}_\text{CLS} = \texttt{CE}(\mathbf{z}^S/\tau_{d}, \mathbf{z}^T/\tau_{d}),
\end{equation}
where $\tau_d$ refers to the temperature parameter, and we simply maintain it as  a constant $\mathbbm{1}$. $\mathbf{z}^S/\mathbf{z}^T$ are the soft label outputs from Student/Teacher network.

\subsection{Training}
For the training, we keep the original VL pre-training objective losses ($\mathcal{L}_\text{VLP}$)~\cite{lu2019vilbert} which consist of: masked language  modeling loss ($\mathcal{L}_\text{MLM}$), where 15\% of the textual tokens are masked and replaced with a special token $\texttt{[MASK]}$ and the VL model is expected to classify these tokens; Image-text (contrastive) matching (ITM) loss ($\mathcal{L}_\text{ITM}$) where the model is expected to predict whether the image-text pair matches. Our final total loss on distillation at the pre-training stage is the combination of the above:
\begin{equation}
    \mathcal{L} = \mathcal{L}_\text{VLP} + \alpha\mathcal{L}_\text{ATT} + \beta\mathcal{L}_\text{HID},
\end{equation}
where $\alpha$ are $\beta$ are the weights of the loss terms. We find that $\mathcal{L}_\text{CLS}$ does not obviously contribute to the pre-training stage so we simply apply it at the fine-tuning distillation stage as:
\begin{equation}
    \mathcal{L} = \mathcal{L}_\text{CE} + \mathcal{L}_\text{CLS} + \alpha\mathcal{L}_\text{ATT} + \beta\mathcal{L}_\text{HID},
\end{equation}
where $\mathcal{L}_\text{CE}$ is the original classification task in the specfic downstream-task. We study the effects of different learning losses in our ablations.

\begin{table*}[t!]
    \centering
    \begingroup
    \setlength{\tabcolsep}{6pt} 
    \renewcommand{\arraystretch}{1.1} 
    {\small
    \begin{tabular}{lccccc|cccc|cc}
    \enspace\enspace\multirow{2}{*}{\textbf{Method}} &  \multirow{2}{*}{\# \textbf{Param}} &  \multirow{2}{*}{\# \textit{I-T} \textbf{Pairs}} &  \multirow{2}{*}{\textbf{Visual Feat.}} & \multirow{2}{*}{\textbf{P. D.}} & \multirow{2}{*}{\textbf{F. D.}} &  \multicolumn{4}{c|}{{ COCO Captioning}} & \multicolumn{2}{c}{VQA}\\ 
    & & & & &  & {B@4} & {M} & {C} & {S} & test-std & test-dev \\
    \hline \rule{0pt}{1.1\normalbaselineskip}
    UVLP~\cite{zhou2020unified}  & $111.7$M & $3$M & ResNeXt101  & \xmark & \xmark & $36.5$ & $28.4$ & $116.9$  & $21.2$  & $70.7$ & $-$\\
    OSCAR$_{\text{B}}$~\cite{li2020oscar}  & $111.7$M & $7$M & R101-F  & \xmark & \xmark & $36.5$   & $30.3$    & $123.7$  & $23.1$ & $73.4$ & $73.2$\\
    MiniVLM~\cite{wang2020minivlm}  & $34.5$M & $7$M & TEE & \xmark & \xmark & $34.3$ & $28.1$ & $116.7$ & $21.3$ & - & -  \\
    MiniVLM~\cite{wang2020minivlm}
 & $34.5$M & $14$M & TEE & \xmark & \xmark & $35.6$ & $28.6$ & $119.8$ & $21.6$ &  $69.4$ & $69.1$ \\ 
    \hline \rule{0pt}{1.05\normalbaselineskip}
    \multirow{4}{*}{DistillV{\small L}M } & \multirow{4}{*}{$34.5$M} & \multirow{4}{*}{$7$M} &
    \multirow{4}{*}{TEE}  & \xmark & \xmark & $34.0$ & $28.0$ & $115.7$ & $21.1$ & 69.0 & 68.8 \\
      &  &  &   & \xmark & \checkmark & \cellcolor{gray!15}$34.5$ & $28.2$ & \cellcolor{gray!20}$117.1$ & $21.5$ & \cellcolor{gray!10}$69.2$ & \cellcolor{gray!10}$69.0$ \\
      &  &  &   & \checkmark & \xmark & \cellcolor{gray!25}$35.2$ & $28.6$ & \cellcolor{gray!25}$120.1$ & $21.9$ & \cellcolor{gray!25}$69.7$ & \cellcolor{gray!25}$69.6$ \\
      &  &  &   & \checkmark & \checkmark & \cellcolor{gray!35}$35.6$ & $28.7$ & \cellcolor{gray!35}$120.8$ & $22.1$ & \cellcolor{gray!25}$69.8$ & \cellcolor{gray!25}$69.6$  \\
    \end{tabular}
    }
    \endgroup
    \vspace{2mm}
    \caption{\small
    \distillvlm distills from stronger VL model (as Teacher), and retains high accuracy on COCO captioning task under different evaluating metrics, regardless of the effect brought by the lightweight visual feature extractor (TEE \textit{v.s.} R101-F). 
    Our model shows competitive results comparing to MiniVLM~\cite{wang2020minivlm}, even only half of the image-text pairs (\# \textit{I-T} Pairs) are available for pre-training. The VL distillation strategy brings consistent improvement in both the pre-training stage (P.D.) and fine-tuning stage (F.D.). All captioning methods are shown with cross-entropy optimization. 
}
\vspace{-4mm}    
\label{tab:main_tbl}
\end{table*}

\section{Experiments}
In this section, we conduct extensive experiments on VL distillation both in pre-training and fine-tuning stages. To evaluate the effectiveness of our proposed distillation schema, we provide results and ablations for the image captioning and VQA tasks.

\subsection{Datasets}
Following~\cite{li2020oscar}, we construct our VL pre-training dataset by combining multiple existing VL datasets. Specifically, we use Conceptual Captions (CC)~\cite{sharma2018conceptual}, SBU captions~\cite{ordonez2011im2text}, training splits of Flicker30k~\cite{plummer2015flickr30k}, GQA~\cite{hudson2019gqa}, COCO Captions~\cite{lin2014microsoft}, and VQA-2.0~\cite{balanced_vqa_v2},~yielding 4 million unique images, and 7 million image-text pairs (VL-7M). Both our Teacher model and \distillvlm are pre-trained on VL-7M and are then transferred to downstream VL tasks: image captioning on COCO Captions and visual question answering on VQA-2.0. 
We follow Karpathy's split\footnote{\url{https://github.com/karpathy/neuraltalk2}} and have $\sim$11k images for training, and $5$k/$5$k images for validation/testing. For the VQA task, we conduct downstream fine-tuning and testing on VQA-2.0 dataset, which consists of $83$k images/$444$k questions for training, $41$k images/$214$k questions for validation. For a fair comparison with previous works, we report results on \texttt{test-std} and \texttt{test-dev} splits via the online evaluation server\footnote{\url{https://visualqa.org/challenge.html}}, and compare ablation results using \texttt{test-dev} split.

\subsection{Implementation Details}
\noindent \textbf{Visual Representation.} Earlier VL pre-training (VLP) works mostly use Faster R-CNN~\cite{anderson2018bottom, ren2015faster} or even advanced architecture~\cite{xie2017aggregated,zhou2020unified} for visual region representation extraction. To obtain visual tokens with more semantics, the object detector for VLP is usually pre-trained on Visual Genome Dataset~\cite{krishna2017visual}, which contains $1,600$ object and 500 attribute categories.
Following MiniVLM~\cite{wang2020minivlm}, we also adopt the EfficientNet~\cite{tan2020efficientdet} based lightweight object detector (TEE) for visual feature extraction. TEE reduces $90\%$ of total inference time and has $91\%$ fewer parameters ($86.9$M for R101-F \vs $7.5$M for TEE). 
Same as MiniVLM, we also pre-train the TEE detector on Object365~\cite{shao2019objects365} and Visual Genome~\cite{krishna2017visual} datasets before the visual representation extraction.
We use R101~\cite{he2016deep} based Faster-RCNN and TEE detected proposals for Teacher's regional visual representation extraction. This guarantees the semantic correspondence of the input tokens between Teacher and Student. 
Prevailing VL pre-training method like~\cite{li2020oscar} shows that applying object tags in VL pre-training contributes to the performances.
During distillation, we use consistent object tags detected by TEE for both the Teacher and Student networks. The lengths for object tags and visual tokens are $15$ and $50$, respectively. \\[-1.6ex]

\noindent \textbf{VL Pre-training\&Distillation.} We use a compact transformer architecture for the VLP and VL distillation. In particularly, we follow~\cite{wang2020minilm,wang2020minivlm} and adopt a $12$-layer transformer with $12$ attention heads and $384$ hidden dimension. For the Teacher model, we use Oscar$_{b}$~\cite{li2020oscar}, a $12$-layer transformer with $12$ attention heads and $768$ hidden size, pre-trained on the VL-7M corpus for 1M steps ($100$ epochs), with learning rate $5e^{-5}$ and batch size $768$, using AdamW optimizer.\footnote{\url{https://github.com/microsoft/Oscar}} Overall, our compact transformer uses the same architecture as MiniVLM~\cite{wang2020minivlm}, and it has $34.5$M learnable parameters and is 70\% less than Oscar$_\text{b}$. 
For VL distillation, we first adapt the Teacher VLM by re-training it using the new visual tokens. Then, we keep the Teacher model frozen without further updating throughout the VL distillation.
In contrast to~\cite{zhou2020unified,li2020oscar}, weights in \distillvlm are randomly initialized without inheriting weights from BERT~\cite{devlin2018bert}. We adopt a learning rate at $2e^{-4}$ with batch size 768 for pre-training/distillation. We report and compare the effect of VL distillation with previous VLP baselines in Table~\ref{tab:main_tbl}. We set $\tau = \tau_d =1$ and $\alpha = 10$, $\beta = 10$. Similar results are observed when using different values. We set the queue size to $4,096$ and further study the effect of different hyper-parameters in ablations. 
\\ [-1.8ex]

\noindent \textbf{Transferring to Downstream Tasks.}
In order to validate the efficacy of our proposed VL distillation schema, we transfer the pre-trained model to VL downstream tasks. Image captioning and VQA task can be formulated as a typical classification task, which enables direct task-specific distillation and comparisons in the downstream. 
We mainly examine them in this work, while the VL distillation is not task-specific and can be extended to other VL tasks as well. We conduct downstream distillation by using the output logit from downstream fine-tuned Teacher as soft-labels ($\mathcal{L}_\text{TASK}$). More details on distillations and ablations for the downstream tasks can be found at Appendix.\\ [-1.8ex]

\noindent \textbf{Image Captioning.} We evaluate our model by transferring it to the image captioning task. 
We fine-tune our model by randomly masking out 15\% of the caption tokens and impose a classification task to predict the masked token id using cross-entropy loss.  Similar to~\cite{devlin2018bert}, we trim and pad textual sentences to the length of $20$. At inference, we recursively feed in [\texttt{MASK}] tokens and predict out captions one after the other with the beam search size at 1. The performance of captioning models is evaluated via BLEU@4~\cite{papineni2002bleu}, METEOR~\cite{denkowski2014meteor}, CIDEr~\cite{vedantam2015cider} and SPICE~\cite{anderson2016spice} metrics. We perform the parameter search in a limited range:  learning rate \{$2e^{-5}$, $5e^{-6}$\} and epochs \{$20$, $30$, $40$\}. \\ [-1.8ex]

\noindent \textbf{VQA.} For the VQA task, the model must select the correct answer from the multi-options list given an image and textual question. We conduct fine-tuning on the VQA-2.0 dataset~\cite{balanced_vqa_v2} and report the accuracy on \texttt{test-std} and \texttt{test-dev} splits. Following~\cite{anderson2018bottom}, we train the VQA model as a $3,129$-way classification task. We perform a light combinatorial parameter search on VQA task within a limited range:  learning rate \{$1e^{-5}$, $5e^{-5}$\} and epochs \{$20$, $40$\}.

\subsection{Results and Analysis}
Table~\ref{tab:main_tbl} summarizes the results of \distillvlm using Oscar$_\text{b}$ as the Teacher model. We list VLP baselines with larger transformer architectures and stronger visual representations in the top lines. In particularly, \distillvlm without VL distillation achieves $34.0$\ BLEU@4 and $115.7$ CIDEr scores with TEE visual representations using VLP~\cite{li2020oscar} (masked language prediction and image-text matching losses).  This is slightly lower than the performance reported by MiniVLM~\cite{wang2020minivlm} pre-trained on VL-7M:  $116.7$ CIDEr score \vs our reproduced $115.7$, which might be caused by the sub-optimal hyper-parameters. 
The apparent performance gaps between larger and smaller VLP models indicate the importance of visual representations so that the VL distillation is desired on small VL architectures. Notably, when equipped with downstream distillation, it performs better on COCO captioning dataset, $1.4$ more on CIDEr, and $0.5$ more on BLEU@4 scores. Downstream distillation on VQA task show marginal improvement: $69.2$ \vs $69.0$. 
We conjecture that this is mainly because the classification distillation on YES/NO or counting type of question does not provide better guidance, that the answers in the VQA task are mostly irrelevant/mutually exclusive.
However, VL distillation in the pre-training stage increases the performances of \distillvlm on both captioning and VQA tasks consistently across all metrics: $\Delta=1.2\%$ at B@4, $4.4$ at CIDEr and $0.7$ higher on VQA test-std split. Compared to its non-distilled counterpart MiniVLM~\cite{wang2020minivlm}, \distillvlm shows better results with only half the size of VL-corpus. To this end, the combination of the VL distillation in both pre-training and fine-tuning stage achieves the best results of \distillvlm, which shows comparable performances with Oscar$_\text{b}$: $120.8$ \vs $123.7$ with $70$\% fewer parameters. To learn more about \distillvlm, we conduct ablations on different designing options and examine the advantages of distillation at different epochs and data usage at Section~\ref{sec:dataefficient}.\\ [-1.8ex]

\begin{table}[t!]
    \centering
    \setlength{\tabcolsep}{4.8pt} 
    \renewcommand{\arraystretch}{1.2} 
    { \small
    \begin{tabular}{ccc|cccc|c}
    \multirow{2}{*}{\textbf{$\mathcal{L}_\text{VLP}$}} & \multirow{2}{*}{\textbf{$\mathcal{L}_\text{ATT}$}} &  \multirow{2}{*}{\textbf{$\mathcal{L}_\text{HID}$}} & \multicolumn{4}{c|}{{ COCO Captioning}} & \multicolumn{1}{c}{{ VQA}}  \\ 
    & & & {B@4} & {M} & {C} & {S} & {test-dev} \\
    \hline 
    \checkmark & \xmark & \xmark & $33.0$ & $27.3$ & $110.6$ & $20.4$ & $68.5$ \\
    \checkmark & \checkmark & \xmark & \cellcolor{gray!0}$32.9$ & $27.5$ & \cellcolor{gray!5}$111.8$ & $20.6$ & \cellcolor{gray!5}$68.9$  \\
    \checkmark & \xmark & \checkmark & \cellcolor{gray!15}$34.0$ & $27.8$ & \cellcolor{gray!15}$114.4$ &  $21.1$ & \cellcolor{gray!15}$69.2$  \\
    \xmark & \checkmark & \checkmark & \cellcolor{gray!10}$33.9$ & $27.8$ & \cellcolor{gray!25}$114.7$ & $21.1$ & \cellcolor{gray!15}$69.2$ \\
    \checkmark & \checkmark & \checkmark & \cellcolor{gray!35}$34.6$  & $27.9$ & \cellcolor{gray!35}$115.6$ & $21.3$ & \cellcolor{gray!25}$69.4$ \\
    \end{tabular}
    }
    \vspace{2mm}
    \caption{\small
    Detailed distillation effects based on attention matrices ($\mathcal{L}_\text{ATT}$), hidden hidden embedding ($\mathcal{L}_\text{HID}$), compared with VL pre-training losses ($\mathcal{L}_\text{VLP}$) at pre-training stage. 
    Results are reported after 20 epochs of pre-training/distillation on 7M Image-Text pairs, then fine-tuned at the downstream (with cross-entropy optimization only).
    }
    \label{tab:att_ablation}
\end{table}

\noindent \textbf{Distillation over Different Losses.} Table~\ref{tab:att_ablation} presents the individual contribution of each distillation loss (attention matrices, hidden embedding) on the basis of the VL pre-training. The experiments for VL Pre-training/Distillation are trained for $20$ epochs using identical hyper-parameters as before. 
From the table, we have the following observations: First, the non-distilled baseline alone reaches $110.6$ CIDEr score for image captioning and $67.2$ accuracy on VQA benchmark (shown in the first line of Table~\ref{tab:att_ablation}). By mimicking the distribution of attention, minor improvements are made, that is $1.2$ for CIDEr and $0.4$ for VQA scores respectively. Similarly, we observe the same trend when combining VLP with hidden embedding distillation. Compared with the VLP baseline, hidden embedding distillation significantly improves the performance under all criteria, demonstrating the efficacy of the alignment schema. In the end, the combination of all the loss terms gives the best performance, confirming that our proposed attention and hidden embedding distillation losses are complementary to each other. We find that using distillation objective alone also produces satisfactory performance, showing that knowledge transfer from distillation is to some extent equivalent to VL pre-training loss.

\begin{table}[t!]
    \centering
    \setlength{\tabcolsep}{3.1pt} 
    \renewcommand{\arraystretch}{1.2} 
    { \small
    \begin{tabular}{l|cccc|c}
   \quad\enspace \multirow{2}{*}{\textbf{Methods}} & \multicolumn{4}{c|}{{ COCO Captioning}} & \multicolumn{1}{c}{{ VQA}}  \\ 
    &  {B@4} & {M} & {C} & {S} & {test-dev} \\
    \hline  \rule{0pt}{1.0\normalbaselineskip}
    VL Pre-training~\cite{li2020oscar} & $33.0$ & $27.3$ & $110.6$ & $20.4$ & $68.5$ \\
    Textual Distill & \cellcolor{gray!10}$34.1$ & $27.7$ & \cellcolor{gray!10}$114.3$ & $20.9$ & \cellcolor{gray!10}$69.0$  \\
    MSE + Layerwise & \cellcolor{gray!10}$34.2$  & $27.8$  & \cellcolor{gray!10}$114.8$ & $21.1$ & \cellcolor{gray!10}$69.2$  \\
    MSE + Last-layer{\color{red} $^{*}$} & \cellcolor{gray!5}$33.3$  & $27.6$ & \cellcolor{gray!5}$112.4$ & $20.7$ & \cellcolor{gray!5}$68.5$ \\
    MSE + Last-layer & \cellcolor{gray!15}$34.3$  & $27.8$ & \cellcolor{gray!15}$115.3$ & $21.2$ & \cellcolor{gray!15}$69.4$ \\
    NCE + Last-layer{\color{red} $^{*}$} & \cellcolor{gray!15}$34.3$ & $27.9$ & \cellcolor{gray!15}$115.4$ & $21.2$  & \cellcolor{gray!12}$69.3$\\
    NCE + Last-layer & \cellcolor{gray!20}$34.6$ & $27.9$ & \cellcolor{gray!25}$115.6$ & $21.3$ & \cellcolor{gray!20}$69.4$\\
    \end{tabular}
    }
    \vspace{2mm}
    \caption{\small
    Ablation of DistillV{\small L}M  using different distillation strategies, \ie, layer-to-layer distillation or last-layer distillation, using mean-square-error distance (MSE) or noise-contrastive (NCE) loss. Textual Distill represents applying the distillation only to the textual tokens without using visual tokens.
    Captioning results are reported after $20$ epochs of training/distillation on VL-7M with cross-entropy optimization. {\color{red} $^{*}$} is the result using mean-pooled token embedding for distillation.
    }
    \label{tab:losses}
\end{table}

\noindent \textbf{Different Distillation Strategies.} \label{sec:method}Table~\ref{tab:losses} shows the results of distillation using different strategies, \ie, layer-to-layer distillation \vs last-layer distillation, and MSE loss \vs NCE loss. We first study the effect of our proposed visual token alignment by applying attention distribution and hidden embedding distillation loss only to the textual token part: \eg, using the ``textual-to-textual'' attention sub-matrices and their corresponded textual token embedding. The second line in Table~\ref{tab:losses} is the result of textual distillation, which shows a slight improvement over the VLP baseline. 
Following previous language distillation works~\cite{jiao2019tinybert}, we also conduct the layer-to-layer attention and hidden distillation between Teacher and Student, and observe inferior performances than the last-layer strategy. Beyond that, the layer-to-layer method can also be severely limited by their architectural structures~\cite{wang2020minilm} (\eg, different number of layers and attention heads). ``NCE + Last-layer'' represents the results of \distillvlm using our proposed contrastive objective function that uses negative samples for the alignment learning. We find that contrastive learning leads to slightly better results than MSE loss. To this end, we study the differences in using token-wise embedding and the mean-pooled layer-wise embedding for contrastive learning and observe that learning with token-wise embedding gives much better results, which is a different observation from~\cite{sun2020contrastive}. However, applying the mean-pooled embedding with NCE loss mitigates this issue and gives on par results with token-wise NCE method (see last two lines in Table~\ref{tab:losses}).
We further provide the ablations of VL distillation for the downstream tasks in the appendix. 

In Table~\ref{tab:neg}, we study the effect of using more negative samples in NCE loss. We observe that increasing the size of the sample queue can steadily contribute to the performances of VL models. Especially, when we only use one negative sample, the model reaches $112.5$ CIDEr score, which aligns with the MSE results ($112.4$ CIDEr score) at Table~\ref{tab:losses}. When increased to $4,096$, the model performs best across all metrics. While continuing to use more negative samples may produce better results, we just set the size of the sample queue as $4,096$ in our experiments. Note that our queue stores the random sample representations from Teacher VLM, which remain consistent throughout the distillation process.
This also implies the feasibility of leveraging in-batch samples for contrastive learning, while the queue design relieves the model from batch-size requirements and allows the use of more negative samples.

\begin{table}[t!]
    \centering
    \setlength{\tabcolsep}{7pt} 
    \renewcommand{\arraystretch}{1.2} 
    { \small
    \begin{tabular}{c|cccc|c}
    \multirow{2}{*}{{\# \textit{Neg}.}} & \multicolumn{4}{c|}{{ COCO Captioning}} & \multicolumn{1}{c}{{ VQA}}  \\ 
    &  {B@4} & {M} & {C} & {S} & {test-dev} \\
    \hline  \rule{0pt}{1.03\normalbaselineskip}
    $1$ & $33.3$  & $27.6$  & $112.5$ & $20.7$ &  $68.5$ \\
    $64$ & \cellcolor{gray!5}$33.6$  & $27.7$  & \cellcolor{gray!5}$112.7$ & $20.9$ &  \cellcolor{gray!5}$68.9$ \\
    $128$ & \cellcolor{gray!10}$33.6$  & $27.7$  & \cellcolor{gray!10}$112.7$ & $20.9$ &  \cellcolor{gray!5}$68.9$ \\
    $512$ & \cellcolor{gray!15}$33.7$  & $27.8$  & \cellcolor{gray!15}$113.3$ & $21.0$ &  \cellcolor{gray!5}$68.8$ \\
    $1,024$ & \cellcolor{gray!20}$34.1$  & $27.9$  & \cellcolor{gray!20}$114.7$ & $21.2$ &  \cellcolor{gray!15}$69.1$ \\
    $4,096$ & \cellcolor{gray!25}$34.3$  & $27.9$  & \cellcolor{gray!25}$115.4$ & $21.2$ & \cellcolor{gray!20}$69.3$  \\
    \end{tabular}
    }
    \vspace{2mm}
    \caption{\small Effect of the number of negative samples for noise contrastive estimation loss. A larger queue size incrementally contributes to the distillation performance. When queue size approaches 1, the NCE loss is approximately the MSE loss with an only positive anchor from the Teacher. All the experiments are trained for 20 epochs using sample queues in different sizes on VL-7M, and then transferred to downstream.}
    \vspace{-2mm}
    \label{tab:neg}
\end{table}

\begin{figure}[t]
    \centering
    \includegraphics[width=.48\textwidth]{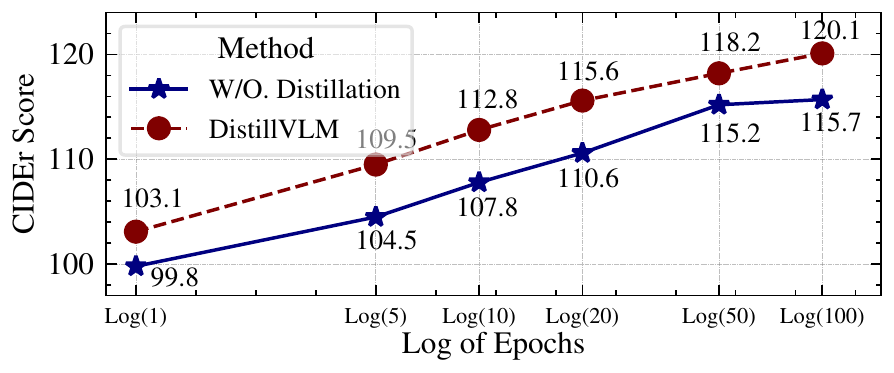} 
    \includegraphics[width=.48\textwidth]{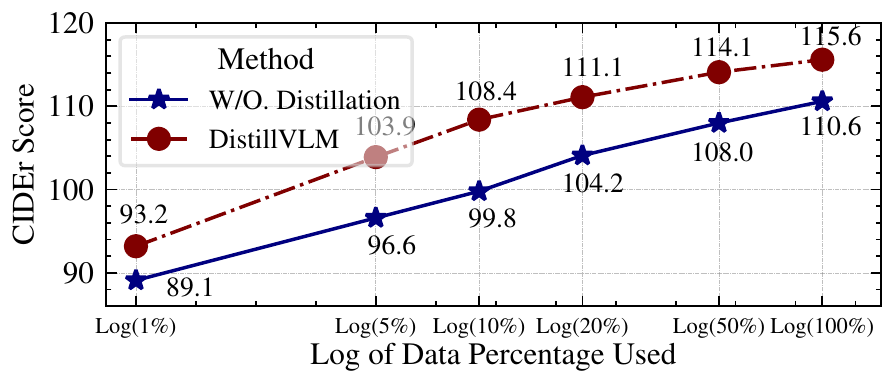} 
    \caption{\small Top: Captioning CIDEr score gain from VL distillation under different epochs ($1$, $10$, $20$, $50$, $100$) in  pre-training/distillation on VL-7M; Bottom: Using $1\%$, $10\%$, $20\%$, $50\%$ and $100\%$ of VL-7M image-text pairs with $20$ epochs of pre-training/distillation. }
    \vspace{-4mm}
    \label{fig:epoch} 
\end{figure}

\vspace{2mm}
\noindent \textbf{Data-efficient VL Distillation.} \label{sec:dataefficient} One critical aspect of VL distillation for real-world application is its ability to efficiently train smaller VL model with limited cost, \ie, with a smaller VL corpus (data scarcity) and less converging epochs (training efficiency). 
To further assess whether VL distillation can cope with these challenges, we perform VL distillation at the pre-training stage when trained with $1$, $5$, $10$, $20$, $50$, $100$ epochs and compare their results with VLP. In addition, as pointed by~\cite{lu202012} that specific partial VL data might contribute more to performances, we propose to conduct VL distillation/pre-training using evenly sampled partial data ($1\%$, $5\%$, $10\%$, $20\%$, $50\%$ and $100\%$ of VL-7M). These also help to verify whether \distillvlm benefits from more converging epochs and more VL data.
Several conclusions can be drawn from the above results. First, VL distillation brings a consistent CIDEr gain across different training epochs. Non-distilled VL pre-training method achieves only $99.8$ CIDEr score with $1$ epoch of training, while \distillvlm reaches $103.1$ (see Figure~\ref{fig:epoch}). Notably, CIDEr score of \distillvlm increases steadily with more training epochs, revealing that VL distillation is more effective than VL pre-training. When it comes to using different percentages of VL data, we also see a similar trend. In the most extreme case, with only $1\%$ of VL-7M corpus available, VL pre-training produces $89.1$ CIDEr score, $4.1$ lower than the VL distillation. With more image-text pairs, VL distillation obviously gives even better results: 
$8.6$ higher for $10\%$, $6.9$ higher for $20\%$ and $5.0\%$ higher for $100\%$.
This shows that regardless of the amount of data available, VL distillation
provides much more effective and informative supervision than the normal pre-training strategy, which leads to better performances.

\section{Conclusion}
We have proposed the first VL distillation, which leverages the knowledge distillation technique to compress large visual-linguistic models. Our experiments confirmed the validity of VL distillation from several aspects: Compared to the non-distilled VL pre-training method, VL distillation not only brings better performances, it is also more data efficient. Our extensive ablations also verified that our VL distillation strategies are simple yet effective.

{\small
\bibliographystyle{ieee}
\bibliography{egbib}
}

\clearpage
\section*{Appendix}

\noindent In this supplementary material, we provide additional details of the VL distillation, which includes: a pseudo-implementation in PyTorch~\cite{paszke2017automatic} style; more results/ablations in the downstream; \distillvlm results transferred to the image retrieval task. To the end, we present qualitative analysis of attention distribution distillation and provide an analysis for VL Distillation and its broader impact for future studies.

\section*{Pseudo-implementation for VL Distillation}
The following section presents the pseudo implementation (see Page~\pageref{fig:pseudocode}) of our proposed VL distillation schema in \textit{Py}Torch style.

\section*{VL Distillation in Image Captioning Task}
We also conduct ablations of VL distillation in the task-specific fine-tuning stage. Table~\ref{tab:downablation} summarizes the fine-tuning results on image captioning task using different distillation losses, whereby weight loss parameters within the range of $[0, 1, 10]$ are searched for. All experiments are pre-trained/distilled for 20 epochs on the VL-7M dataset. In contrast to the results in pre-training stage VL distillation, we find that the $\mathcal{L}_\text{HID}$ and $\mathcal{L}_\text{att}$ do not contribute to the performances for downstream VL distillation. Instead, using just the $\mathcal{L}_\text{CLS}$ obviously improves the results: the CIDEr score is improved from $115.6$ to $116.3$.

\section*{Transferring to Image-text Retrieval Task}
In addition to the previously mentioned image captioning task and VQA task, \distillvlm is not VL task specific and can be extended to other tasks as well. In this section, we study whether the advantages from pre-training stage VL distillation extend to the image-text retrieval task on COCO dataset. In particular, the image-text retrieval task aims to retrieve the target image or text given a query text or image. We set up the image-text retrieval task on COCO dataset, and evaluate the performance of model using top-\textit{K} retrieval accuracy on both 1k and 5k test splits. To transfer and train the VL model for the image-text retrieval task, we use the first $\texttt{[CLS]}$ token from transformer in a binary classification task, where the classifier predicts whether the image and text match or not. To construct the negative tuple, irrelevant texts are sampled as non-matching cases. We fine-tune the model with learning rate at $2e-5$ and train it for $30$ epochs. The results are shown in Table~\ref{tab:retrieval}.

\begin{table}[t!]
    \centering
    \setlength{\tabcolsep}{3.8pt} 
    \renewcommand{\arraystretch}{1.2} 
    { \small
    \begin{tabular}{ccc|cccc}
    \multirow{2}{*}{\textbf{$\mathcal{L}_\text{CLS}$}} & \multirow{2}{*}{\textbf{$\mathcal{L}_\text{ATT}$}} &  \multirow{2}{*}{\textbf{$\mathcal{L}_\text{HID}$}} & \multicolumn{4}{c}{{ COCO Captioning}}   \\ 
    & & & {B@4} & {M} & {C} & {S} \\
    \hline 
    $0$ & $0$ & $0$ & $34.6$ & $27.9$ & $115.6$ & $21.3$ \\
    $1$ & $0$ & $0$ & $34.6$ & $28.1$ & $116.3$ & $21.3$\\
    $1$ & $1$ & $0$ & $34.6$ & $28.1$ & $116.3$ & $21.3$ \\
    $1$ & $10$ & $0$ & $34.5$ & $28.0$ & $116.1$ & $21.3$ \\
    $1$ & $0$ & $1$ & $34.5$ & $28.0$ & $116.0$ & $21.3$ \\
    $1$ & $0$ & $10$ & $34.3$ & $28.0$ & $115.2$ & $21.2$ \\
    $10$ & $10$ & $10$ &  $34.6$ & $28.1$ & $116.4$ & $21.3$\\

    \end{tabular}
    }
    \vspace{2mm}
    \caption{\small
    Detailed distillation effects based on classification logit ($\mathcal{L}_\text{CLS}$), attention matrices ($\mathcal{L}_\text{ATT}$), hidden embedding ($\mathcal{L}_\text{HID}$) with various weights ($\alpha$, $\beta$), compared with non-distilled strategy at fine-tuning stage. The first line shows the fine-tuning result without any distillation.
    }
    \label{tab:downablation}
\end{table}

\begin{table*}[t!]
\begin{center}
\renewcommand{\arraystretch}{1.2} 
{\small
\begin{tabular}{lcc@{~~}c@{~~}cc@{~~}c@{~~}cc@{~~}c@{~~}cc@{~~}c@{~~}c}
\toprule
\hskip15pt \multirow{4}{*}{\textbf{Method}}  & & \multicolumn{6}{c}{\textbf{1K Test Set}} & \multicolumn{6}{c}{\textbf{5K Test Set}}     \\ 
\cmidrule(lr){3-8} \cmidrule(lr){9-14} & & \multicolumn{3}{c}{Text Retrieval}  & \multicolumn{3}{c}{Image Retrieval}   & \multicolumn{3}{c}{Text Retrieval}   & \multicolumn{3}{c}{Image Retrieval}  \\ [2px]
\cmidrule(lr){3-5} \cmidrule(lr){6-8} \cmidrule(lr){9-11}\cmidrule(lr){12-14} &    & R@1  & R@5  & R@10 & R@1 & R@5 & R@10  & R@1 & R@5 & R@10 & R@1 & R@5 & R@10  \\ 
\hline
PixelBERT~\cite{huang2020pixel} & \xmark   & $77.8$  & $95.4$ & $98.2$ & $64.1$ & $91.0$ & $96.2$ & $53.4$ & $80.4$ & $88.5$ & $41.1$ & $69.7$ & $80.5$ \\  
Unicoder-VL~\cite{li2020unicoder}& \xmark   & $84.3$  & $97.3$ & $99.3$ & $69.7$ & $93.5$ & $97.2$ & $62.3$ & $87.1$ & $92.8$ & $46.7$ & $76.0$ & $85.3$ \\
OSCAR$-\text{base}$~\cite{li2020oscar}& \xmark  & $88.4$  & $99.1$ & $99.8$ & $75.7$ & $95.2$ & $98.3$ & $70.0$ & $91.1$ & $95.5$ & $54.0$ & $80.8$ & $88.5$ \\ 
MiniVLM~\cite{wang2020minivlm} & \xmark &$81.1$  &$96.1$ & $99.2$   &$68.5$ &$93.0$  & $97.1$  &$58.8$  & $85.1$ & $91.7$ &$45.0$ &$74.1$   & $84.0$ \\ 
\hline
Baseline  & \xmark & $77.3$ & $95.0$  & $98.5$  & $64.6$ & $92.0$   & $96.7$ & $54.4$ & $81.4$  & $89.3$ & $41.3$ & $71.2 $ & $81.9$ \\
DistillVLM  & \checkmark & $80.0$ & $95.5$ & $98.5$ & $68.3$ & $92.3$ & $96.9$ & $58.3$ & $84.1$ & $91.3$ & $43.9$ & $73.7$ & $83.3$\\ [-5pt]
    \quad\enspace $\Delta$ & &  {\scriptsize$\mathcolor{darkgreen}{\textbf{+2.7}}$} &   {\scriptsize$\mathcolor{darkgreen}{\textbf{+0.5}}$} &   {\scriptsize{+0.0}} &   {\scriptsize$\mathcolor{darkgreen}{\textbf{+3.7}}$} &  {\scriptsize$\mathcolor{darkgreen}{\textbf{+0.3}}$} & {\scriptsize$\mathcolor{darkgreen}{\textbf{+0.2}}$} &   {\scriptsize$\mathcolor{darkgreen}{\textbf{+3.9}}$} &   {\scriptsize$\mathcolor{darkgreen}{\textbf{+2.7}}$} &   {\scriptsize$\mathcolor{darkgreen}{\textbf{+2.0}}$} &   {\scriptsize$\mathcolor{darkgreen}{\textbf{+2.6}}$} &   {\scriptsize$\mathcolor{darkgreen}{\textbf{+2.5}}$} &   {\scriptsize$\mathcolor{darkgreen}{\textbf{+1.4}}$}\\

\bottomrule
\end{tabular}
\vspace{1mm}
\caption{Results of \distillvlm with and without VL distillation transferring to Image-Text Retrieval task  on COCO dataset.}
\label{tab:retrieval}
}
\end{center}

\end{table*}

\begin{table}[t!]
    \centering
    \setlength{\tabcolsep}{4.5pt} 
    \renewcommand{\arraystretch}{1.2} 
    { \small
    \begin{tabular}{c|cccc}
    \multirow{2}{*}{\textbf{Method}} & \multicolumn{4}{c}{{COCO Captioning }} \\ 
    & {B@4} & {M} & {C} & {S} \\
    \hline
    VL Pre-training & $33.0$ & $27.3$ & $110.6$ & $20.4$ \\
    VL Distill. w/o. Adaptation & $33.3$ & $27.5$ & $112.3$ & $20.6$ \\
    VL Distill. with Adaptation.$^{\color{red} *}$ &  $34.4$ & $27.9$ & $115.0$ & $21.0$\\
    VL Distill. with Adaptation & $34.6$ & $27.9$ & $115.6$ & $21.3$ \\

    \end{tabular}
    }
    \vspace{2mm}
    \caption{\small
    Effect of VL distillation with and without Teacher VLM adaptation in pre-training stage. First row in the table shows the captioning performances using VL pre-training without VL distillation. $^{\color{red} *}$ denotes the VL distillation results without re-training the Teacher VLM.
    }
    \label{tab:vlpadapt1}
\end{table}

\begin{table}[t!]
    \centering
    \setlength{\tabcolsep}{5.pt} 
    \renewcommand{\arraystretch}{1.2} 
    { \small
    \begin{tabular}{c|cccc}
    \multirow{2}{*}{\textbf{Method}} & \multicolumn{4}{c}{{COCO Captioning}} \\
    & {B@4} & {M} & {C} & {S} \\
    \hline 
    VL Pre-training & $33.0$ & $27.3$ & $110.6$ & $20.4$ \\
    VL Distill. w/o. Adaptation & $33.2$   & $ 27.5$ & $110.7$  & $20.3$\\
    VL Distill. with Adaptation & $33.5$ & $27.5$ & $111.6$ & $20.6$\\
    \end{tabular}
    }
    \vspace{2mm}
    \caption{\small
    Effect of VL distillation with and without Teacher VLM adaptation in fine-tuning stage. First row in the table shows the COCO captioning results without any downstream distillation.
    }
    \label{tab:vlpadapt2}
\end{table}

We just examine the effect of VL distillation in the pre-training stage. 
DistillVLM without VL distillation is notably worse than MiniVLM: $81.1$ \vs $77.3$ accuracy of R@$1$ on 1$K$ Test Set,  since less training data is used. Equipped with pre-training stage VL distillation, it is significantly improved to $80.0$ and $58.3$ accuracy of R@$1$ on $1$K/$5$K Test Set respectively, which are $2.7$ and $4.0$ higher. Compared to its counterpart MiniVLM~\cite{wang2020minivlm} and other state-of-the-art methods, \distillvlm achieves on par performances, while MiniVLM is trained on 14M image-text pairs, which is twice the size of our pre-training VL corpus. Other methods all use a larger transformer model or stronger visual representations.

\section*{Discussion on Teacher VLM Adaptation}

\begin{table*}[t!]
    \centering
    \setlength{\tabcolsep}{7pt} 
    \renewcommand{\arraystretch}{1.2} 
    { \small
    \begin{tabular}{ccccc|ccccc}
    \multirow{2}{*}{{\# \textit{Epochs}}} & \multicolumn{4}{c|}{{ COCO Captioning}} & \multirow{2}{*}{{\% \textit{Data}}} & \multicolumn{4}{c}{{ COCO Captioning}} \\ 
    &  {B@4} & {M} & {C} & {S}  & &   {B@4} & {M} & {C} & {S} \\ 
    \hline
    \rowcolor{gray!15}\multirow{2}{*}{{ {1}}} & $31.3$ & $25.6$  & $99.8$ & $19.0$ & \multirow{2}{*}{{ {1\%}}} & $27.9$  & $24.2$  & $89.1$ & $17.2$ \\
    & $31.3$  & $26.2$  & $103.3$ & $19.4$ & & $29.2$  & $24.6$  & $93.2$ & $17.8$ \\
    \hline
    \rowcolor{gray!15}
    \multirow{2}{*}{{ {5}}} & $31.3$  & $26.4$  & $104.5$ & $19.6$ & \multirow{2}{*}{{ {5\%}}} & $29.5$  & $25.1$  & $96.6$ & $18.4$ \\
    & $32.8$  & $27.3$  & $109.5$ & $20.5$ & & $31.6$  & $26.3$  & $103.9$ & $19.3$ \\
    \hline
    \rowcolor{gray!15}
    \multirow{2}{*}{{ {10}}} & $32.2$  & $26.8$  & $107.8$ & $20.2$ & \multirow{2}{*}{{ {10\%}}} & $30.2$  & $25.7$  & $99.8$ & $18.9$ \\
    & $33.9$  & $27.5$  & $112.8$ & $20.6$ & & $32.8$  & $27.0$  & $108.4$ & $20.0$ \\
    \hline
    \rowcolor{gray!15}
    \multirow{2}{*}{{ {20}}} & $33.0$  & $27.3$  & $110.6$ & $20.4$ & \multirow{2}{*}{{ {20\%}}} & $31.2$  & $27.3$  & $104.1$ & $19.6$ \\
    & $34.6$ & $27.9$ & $115.6$ & $21.3$ & & $33.4$  & $27.3$  & $111.1$ & $20.5$ \\
    \hline
    \rowcolor{gray!15}
    \multirow{2}{*}{{ {50}}} & $34.3$  & $27.8$  & $115.2$ & $21.1$ & \multirow{2}{*}{{ {50\%}}} & $32.2$  & $26.9$  & $108.5$ & $20.1$ \\
    & $35.1$  & $28.2$  & $118.2$ & $21.5$ & & $34.0$  & $27.7$  & $114.1$ & $21.2$ \\
    \hline
    \rowcolor{gray!15}
    \multirow{2}{*}{{ {100}}} & $34.0$  & $28.0$  & $115.7$ & $21.1$ & \multirow{2}{*}{{ {100\%}}} & $33.0$  & $27.3$  & $110.6$ & $20.4$ \\
    & $35.2$ & $28.6$ & $120.1$  & $21.9$ & & $34.6$  & $27.9$  & $115.6$ & $21.3$ \\
    \end{tabular}
    }
    \vspace{4mm}
    \caption{\small Detailed image captioning results for data efficient VL distillation. The rows in gray show results using non-distilled VL pre-training method, and the following rows show results with VL distillation.}
    \label{tab:efficient}
\end{table*}

Unlike the previous knowledge distillation where the Teacher model is usually not re-trained, Teacher VLM in our work is adapted using the new visual tokens. This is to ensure that the use of new Teacher visual tokens does not have a negative impact to the Teacher. Nevertheless, it is also viable to use Teacher VLM directly without adaptation. We compare the performances using Teacher VLM with and without adaptation in Table~\ref{tab:vlpadapt1} and~\ref{tab:vlpadapt2} in both pre-training and fine-tuning stages. 

In particular, we first conduct VL distillation in pre-training stage without using the adapted Teacher VLM, and use the old visual tokens as in~\cite{li2020oscar} without re-extracting them using object proposals from the lightweight detector. As there is no semantic correspondence across Teacher and Student's tokens, we just adopt the classification distillation on the masked tokens' logit. The second line of Table~\ref{tab:vlpadapt1} gives the results (VL Distill. w/o. Adaptation), where we observe a slight improvement on captioning performances: $112.3$ \vs $110.6$. Furthermore, we apply the visual tokens extracted according to the Student detector's object proposals, which guarantees the semantic correspondence across Teacher and Student tokens. With that, we conduct the VL distillation using the our proposed losses ($\mathcal{L}_\text{ATT}$ and $\mathcal{L}_\text{HID}$), and find significant improvement: CIDEr score increases from $110.6$ to $115.0$ (as shown in third line of Table~\ref{tab:vlpadapt1}). Finally, by re-training the Teacher VLM using the new extracted visual tokens will lead to more improvement: $115.6$ \vs $115.0$. This confirms that, even without re-training the Teacher VLM, VL distillation still gives satisfactory results.

We conduct similar investigations in the downstream and show results in Table~\ref{tab:vlpadapt2}. In particular, downstream VL Distillation without adaptation gives minor improvement ($0.2$ and $0.1$ for BLEU@$4$ and CIDEr scores), while using the adapted Teacher VLM leads to obvious better results ($0.5$ and $1.0$ higher on BLEU@$4$ and CIDEr score). We only apply the classification distillation loss in the downstream experiments.

\section*{Detailed Results for Data Efficient VL Distillation }
Table~\ref{tab:efficient} summarizes more detailed results of data efficient VL distillation on image captioning tasks.  In particular, we compare the effectiveness of  VL distillation in pre-training stage using different converging epochs $[1, 5, 10, 20, 50, 100]$ and fraction data $[1\%, 5\%, 10\%, 20\%, 50\%, 100\%]$.  VL distillation comprehensively improves the results on all metrics, regardless of the amount of data available or different training epochs.

\section*{Effect of Attention Distribution Distillation}
\label{sec:adapt}
\begin{figure*}
\centering
\begin{subfigure}{.5\textwidth}
  \centering
    \includegraphics[width=.9\textwidth]{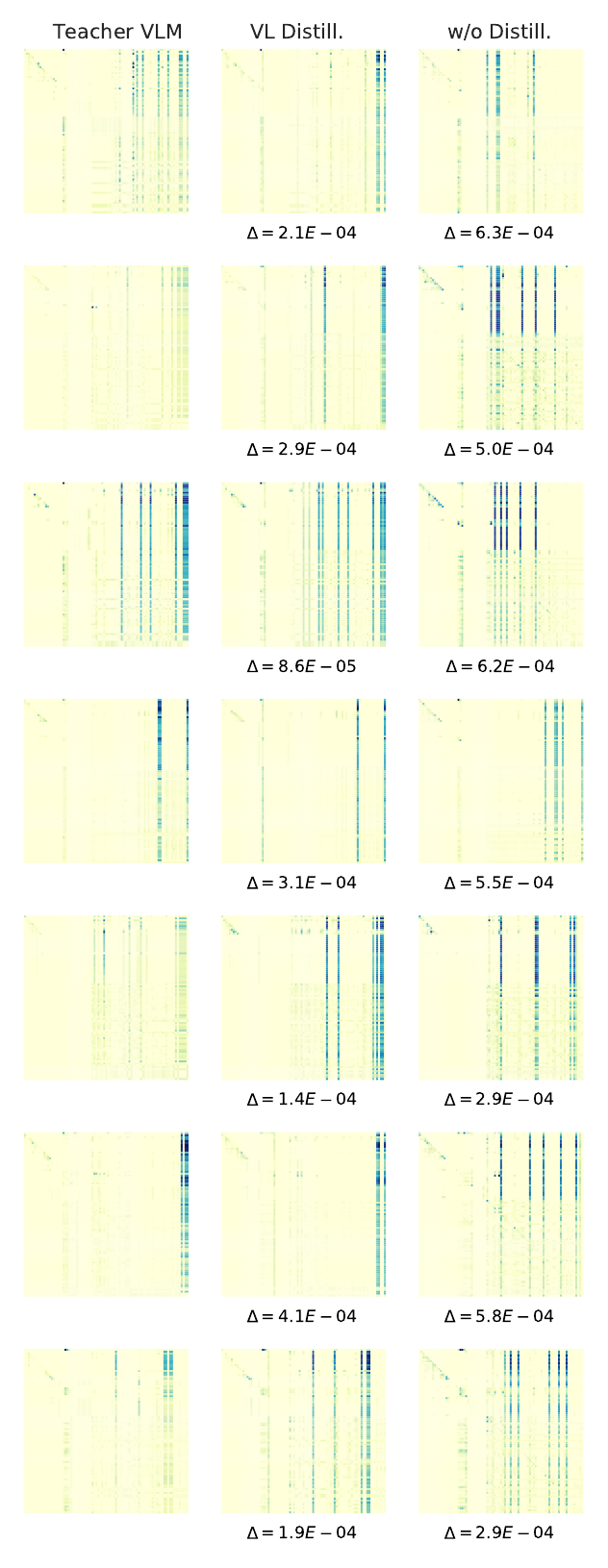} 
  \label{fig:sub1}
\end{subfigure}%
\begin{subfigure}{.5\textwidth}
  \centering
    \includegraphics[width=.9\textwidth]{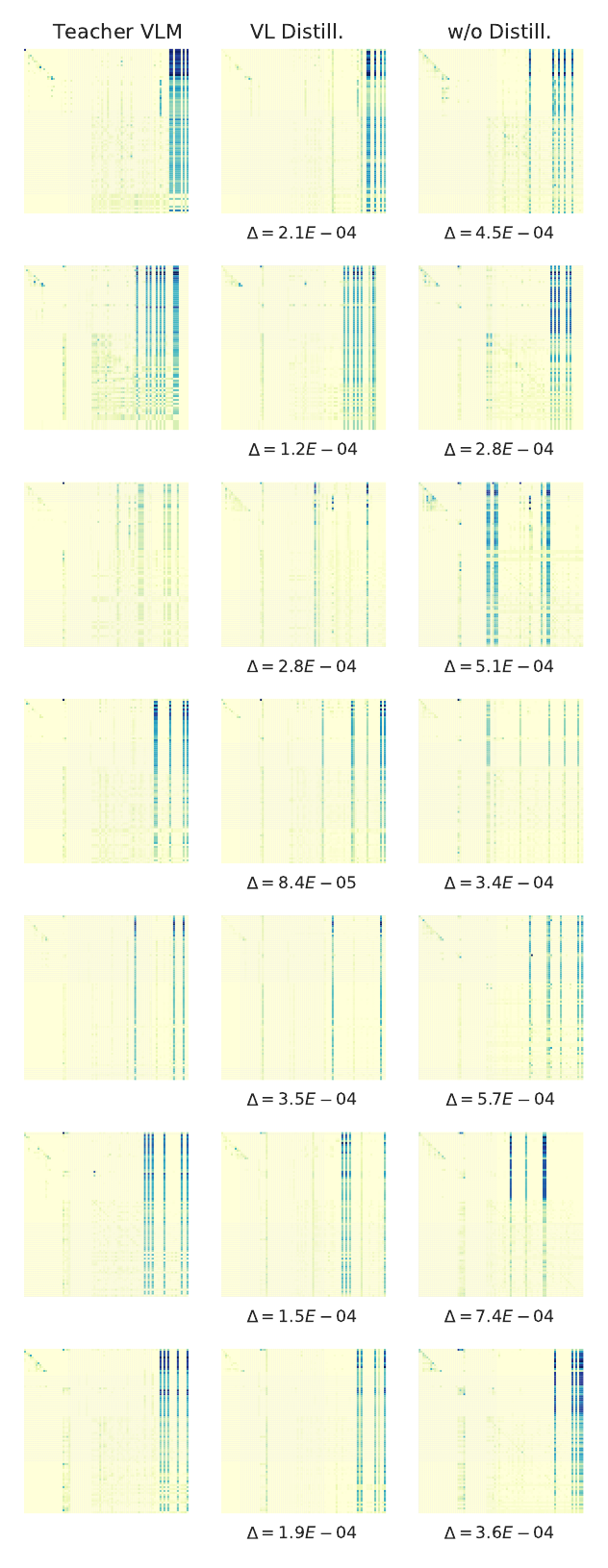} 
  \label{fig:sub2}
\end{subfigure}

\caption{\small Qualitative effect of attention distribution distillation. The 3 neighbouring attention maps are the averaged attention map from the last transformer block of: Teacher VLM, \distillvlm, and non-distilled VL model. The number at bottom is the $l2$ distance of the attention map with its corresponded Teacher attention map.}
\label{fig:att}
\end{figure*}

In Figure~\ref{fig:att}, we show qualitative effect of attention distribution distillation where we select random samples from VL corpus and retrieve their last transformer block attention maps from the Teacher VLM, \distillvlm and the non-distilled counterpart of \distillvlm. To illustrate this, the attention maps are averaged over 12 self-attention heads. Based on the examples, we observe that the attention maps after VL distillation exhibit higher similarities to the Teacher VLM's attention maps. This confirms that our attention distribution distillation successfully mimic the Teacher VLM's attention distribution to some extent.

\begin{figure*}[h]
    \noindent\begin{minipage}{2.14\columnwidth}
    {\centering
        \lstset{
            language={python},
            backgroundcolor=\color{codegray},  
            commentstyle=\color{darkgreen}\ttfamily,
            commentstyle=\color{darkgreen},
            keywordstyle=\color{magenta},
            numberstyle=\tiny\color{maroon},
            stringstyle=\color{codepurple},
            basicstyle=\ttfamily\footnotesize,
            breakatwhitespace=false,         
            breaklines=true,                 
            captionpos=b,                    
            keepspaces=true,                 
            numbers=left,                    
            numbersep=5pt,                  
            showspaces=false,                
            showstringspaces=false,
            showtabs=false,                  
            tabsize=2,
            moredelim=**[is][\color{maroon}]{<}{>},
            moredelim=**[is][\color{blue}]{--}{--},
            moredelim=**[is][\color{maroon}]{```}{```},
            moredelim=**[is][\color{p}]{---}{---},
        }
        \begin{lstlisting}
        ```--Q:-- Maintaining queue of random history representations: (---K--- X ---DT---)
        --teacher_model, student_model:-- Large/small Teacher/Student VL model. 
        --FC--: Linear layer for dimension transformation of student hidden representation.
        --temp:-- Temperatures of the NCE loss.
        --lamda, beta--: Weights for loss terms.
        ```
        
        # activate evaluation mode for Teacher to freeze BN and updation.
        with torch.no_grad():
            # t_atts: 1 X N X N, t_hids: 1 X N X DT
            t_atts, t_hids = --teacher_model--(**teacher_inputs)
        
        # s_atts: 1 X N X N, s_hids: 1 X N X DS
        mlm_loss, itm_loss, s_atts, s_hids = --student_model--(**student_inputs)
        
        # dimension transformation on student hidden, s_hids: 1 X N X DT
        s_hids = --FC--(s_hids)
        
        # attention distillation loss
        att_loss = <mse_loss>(s_atts, t_atts)

        # NCE hidden states distillation loss
        
        # l2-norm the hidden representation
        t_hids, s_hids = <L2_norm>(s_hids), <L2_norm>(t_hids)

        # positive logit: Nx1
        l_pos = torch.<einsum>('nd,nd->n',  [s_hids.squeeze(0), t_hids.squeeze(0)])

        # negative logit: NxK
        l_neg = torch.<einsum>('nd,dk->nk', [s_hids.squeeze(0), Q.<detach>().<T>()])

        # logit: Nx(1+K)
        logits = torch.<cat>([l_pos, l_neg], dim=1)

        # apply the temperature for NCE loss
        logits /= --temp--

        # NCE label: N
        labels = torch.<zeros>(logits.shape[0], dtype=torch.long).cuda()

        nce_loss = <ce_loss>(logits, labels)

        # update the queue with current teacher hidden states.
        <dequeue_and_enqueue>(t_hids, Q)
        
        # aggregated losses
        loss = mlm_loss + itm_loss + --lamda-- * att_loss + --beta-- * nce_loss
        \end{lstlisting}
    }
    \end{minipage}
    \vspace{-4mm}
\caption{Our VL distillation consists of the attention distribution and hidden embedding mimicking when batch size $= 1$ with $N$ textual/visual tokens. We assume \texttt{[PAD]} tokens are excluded from input tokens before the distillation. }
\label{fig:pseudocode}
\end{figure*}

\clearpage

\clearpage

\end{document}